\documentclass[journal]{IEEEtran}
\pdfoutput=1
\ifCLASSINFOpdf
   \usepackage[pdftex]{graphicx}
\else
\fi
\DeclareUnicodeCharacter{0301}{\'{f}}
\usepackage{diagbox}
\usepackage{amsmath}
\usepackage{amsfonts}
\usepackage{mathtools}
\usepackage[para]{threeparttablex}
\usepackage{booktabs}
\usepackage{rotating}
\usepackage{pifont}
\newcommand{\cmark}{\ding{51}}%
\newcommand{\xmark}{\ding{55}}%
\usepackage[style=ieee, citestyle=numeric-comp, backend=biber, maxbibnames= 3]{biblatex}
\renewbibmacro*{volume+number+eid}{%
  \printfield{volume}%
  \setunit*{\addnbspace}
  \printfield{number}%
  \setunit{\addcomma\space}%
  \printfield{eid}}

\AtBeginBibliography{\small}
\usepackage{xcolor}
\usepackage{soul}
\usepackage[normalem]{ulem}
\everypar=\expandafter{\the\everypar\loosness=-1 }
\DeclareUnicodeCharacter{2212}{-}
\begin{document}
%

\title{Enhancing Multivariate Time Series Classifiers through Self-Attention and Relative Positioning Infusion}
%
%
%

\author{Mehryar~Abbasi,~\IEEEmembership{Student Member, IEEE,}
        Parvaneh~Saeedi,~\IEEEmembership{Member,~IEEE}


}
\maketitle

\begin{abstract}
Time Series Classification (TSC) is an important and challenging task for many visual computing applications. Despite the extensive range of methods developed for TSC, relatively few utilized Deep Neural Networks (DNNs). In this paper, we propose two novel attention blocks (Global Temporal Attention and Temporal Pseudo-Gaussian augmented Self-Attention) that can enhance deep learning-based TSC approaches, even when such approaches are designed and optimized for a specific dataset or task. We validate this claim by evaluating multiple state-of-the-art deep learning-based TSC models on the University of East Anglia (UEA) benchmark, a standardized collection of 30 Multivariate Time Series Classification (MTSC) datasets. We show that adding the proposed attention blocks improves base models' average accuracy by up to 3.6\%. Additionally, the proposed TPS block uses a new injection module to include the relative positional information in transformers. As a standalone unit with less computational complexity, it enables TPS to perform better than most of the state-of-the-art DNN-based TSC methods. The source codes for our experimental setups and proposed attention blocks are made publicly available\footnote{https://github.com/mehryar72/TimeSeriesClassification-TPS}.

\end{abstract}

\begin{IEEEkeywords}
Multivariate Time Series Classification, Temporal Attention, Positional Information, Time Series Analysis.
\end{IEEEkeywords}

%
\IEEEpeerreviewmaketitle

\section{Introduction}
\IEEEPARstart{T}{\lowercase{ime}} series data is a set of data points representing qualitative or quantitative information over a time interval. The significance of any series depends on the order and timing of its data points. TSC is the task of classifying time series data based on their attributes over the time they are collected. The process is called Univariate Time Series Classification (UTSC) if the data points only have a single dimension. In contrast, classifying time series data with multidimensional data points is called Multivariate Time Series Classification (MTSC). Many real-world tasks, such as human activity recognition, machine condition monitoring~\cite{TJNN_tsc_mcm}, electrocardiogram (ECG)~\cite{TJNN_tsc_ecg} and electroencephalography (EEG) classification~\cite{TJNN_tsc_eeg}, facial action unit classification~\cite{TJNN_tsc_fac,sim}, and more~\cite{TJNN_tsc_cyc,bagnall2018uea} could be categorized as a TSC problem. As a result, TSC is an active research subject~\cite{bagnall2018uea}. Deep learning-based TSC algorithms tend to have simpler implementations, shorter training periods, fewer computational complexities, and more scalability than traditional methods. However, they are outnumbered by traditional methods because of their lower classification accuracies. Specifically, only one method has demonstrated competitive performance compared to the traditional methods~\cite{fawaz2020inceptiontime,2021hive}. One noticeable issue with deep learning-based TSC algorithms is their incapacity to generalize for different applications. For instance, a model may have superior performance for one task but inferior performance for another~\cite{rank1}. It has, therefore, been difficult to develop a general deep learning-based solution suitable for all TSC applications. There seems to be room for extending and improving deep learning-based TSC methods to be more accurate and general for different tasks.

This paper presents two deep learning-based modules that can be directly integrated into any Deep Neural Network (DNN) TSC model to improve performance. We introduce two new attention-based processing blocks called Global Temporal Attention (GTA), and Temporal Pseudo-gaussian augmented Self-attention (TPS). While the application format of these two blocks is different. They both improve any model regardless of the task. We present the improvement gained by the addition of the proposed blocks to four well-known and popular deep learning-based TSC models (FCN~\cite{wang2017time}, ResNet~\cite{wang2017time}, InceptionTime~\cite{fawaz2020inceptiontime}). The performance of these baseline models before and after adding the suggested attention modules are compared on UEA~\cite{bagnall2018uea} benchmark dataset collection.

Our main contribution here is the introduction of GTA and TPS blocks. These two blocks use attention to underline informative temporal points and data sections unique to each class. They make the learning process of distinguishing between classes easier for any DNN-based TSC.

\section{Related Work}
We categorize related works into three categories. The first category is focused on traditional (non-deep learning) TSC methods. The second category is Deep Neural Network (DNN) TSCs. The proposed work in this paper is among this group of works. The last category is the group that focuses on injecting position information into Transformer models. These works are primarily concerned with Natural Language Processing (NLP) tasks. However, according to ~\cite{dufter2022position}, the way they process the position information is similar to the proposed TPS block and, therefore, worthy of review in this section.

\subsection{Traditional TSC methods}
The most common method of this group is the “NN-DTW” approach, which is composed of the nearest neighbor (NN) classifier with Dynamic Time Warping (DTW) distance metric~\cite{lines2015time}. Recently, the field has been dominated by highly complex classifiers such as Shapelet Transform~\cite{bagnall2017great}, BOSS~\cite{bagnall2017great}, and HIVE-COTE~\cite{lines2018time,2021hive} (ranked as the most accurate classifiers on UCR archive\cite{bagnall2017great,2021hive}). The most recent classifiers can be divided into two categories of simple and complex methods. Complex classifiers are cumbersome, memory intensive, and difficult to train, making them highly unscalable. In contrast, simple methods are faster and relatively easier to train but usually less accurate. 
\subsubsection{\textbf{Complex Traditional TSC}}
In this subcategory, we will review several famous, highly complex, computation-heavy, and accurate traditional TSC classifiers.

Shapelet Transform classifier identifies discriminative subseries (i.e., shapelets)~\cite{bagnall2015time} unique to each class. For each input, shapelets are slid along the time dimension. The distance between the shapelet and the sample at each time is used to form a new array (transformation). The classification step is performed using transformed arrays. The Shapelet transform is one of the most computationally complex methods that escalates higher depending on the number of training samples and the time series’ length. Bag-of-SFA-Symbols (BOSS) is a dictionary-based ensemble classifier model that transforms the frequency of the patterns’ occurrence into a new format. Word extraction for time series classification (WEASEL)~\cite{schafer2017fast} applies static feature selection on the output of a dynamically-sized sliding window feature extractor. It achieves higher accuracy than BOSS but has similar training complexities and high memory usage. Collective Of Transformation-based Ensembles (COTE)~\cite{bagnall2017great} is a large ensemble of 35 different classifiers, including BOSS and Shapelet Transform. 

Hierarchical Vote Collective of Transformation-based Ensembles (HIVE-COTE)~\cite{lines2018time} extends the COTE system to include a new hierarchical structure with probabilistic voting and two new classifiers. It includes two new functions to project time-series arrays into new feature spaces. Although HIVE-COTE has become one of the leading TSC algorithms, it is a highly complex algorithm with many hyper-parameters, requiring high memory usage and computational resources. Temporal Dictionary Ensemble (TDE)~\cite{middlehurst_temporal_2021} is a new ensemble of dictionary-based classifiers similar to BOSS. It combines design features from multiple methods, such as BOSS and WEASEL, to create a more accurate approach.~\cite{middlehurst_temporal_2021} showed that HIVE-COTE’s accuracy can significantly increase if BOSS is replaced with TDE. Hive-cote 2.0~\cite{2021hive} is an upgraded version of HIVE-COTE, which includes comprehensive changes through the compilation of scattered works such as TDE, which significantly improved its accuracy.

Even though complex traditional TSC methods are highly accurate, their complexity makes them unscalable and less practical. Training for some of these algorithms might take weeks to complete. Therefore, there exists a demand for scalable and less memory-intensive methods.

\subsubsection{\textbf{Simple Traditional TSC}}
Simple traditional TSCs are a set of algorithms designed to be faster, less complex, less memory intensive, and easier to train than complex traditional TSC methods.
This subsection reviews more scalable traditional TSC methods such as Proximity Forest~\cite{lucas2019proximity}, TS-CHIEF~\cite{shifaz2019ts}, and ROCKET~\cite{dempster2020rocket}.

The Proximity Forest is an elastic ensemble of proximity decision trees, where the samples are compared against branch exemplars with a randomly chosen distance metric for each node~\cite{lucas2019proximity}. The Time Series Combination of Heterogeneous and Integrated Embedding Forest (TS-CHIEF) extends the Proximity Forest by combining interval-based and dictionary-based branching~\cite{shifaz2019ts}. Although these methods are more scalable, they are still highly complicated, with training complexities that are quadratic in time series length. RandOm Convolutional KErnel Transform (ROCKET) was proposed as a high-speed, high-accuracy method for TSC~\cite{dempster2020rocket}. ROCKET requires only 5 minutes of training on the longest UCR archive time series~\cite{dempster2020rocket}. For comparison, TS-CHIEF requires four days for training on that same dataset~\cite{dempster2020rocket}. ROCKET uses many random convolution kernels in combination with a ridge regression classifier. A notable limitation of ROCKET is its requirement for an extensive collection of diverse data to establish a general feature space. Moreover, it shows a lower performance on unseen datasets. MiniRocket~\cite{dempster_minirocket_2021} is a faster version of ROCKET that uses a deterministic approach toward selecting Kernels and their specifications. MultiRocket~\cite{tan_multirocket_2022} increases the accuracy of MiniRocket by generating more diverse features and utilizing pooling and transform operations.

Although simple traditional TSCs seem more scalable and less computationally expensive than the complex traditional TSCs, they are nonetheless unscalable and computationally expensive if compared to the non-traditional TSC methods. In retrospect, deep learning-based TSC classifiers are much easier to train and considerably more scalable than traditional classifiers (both simple and complex categories).
\subsection{DNN-based TSC}
DNN methods’ simpler implementations, shorter training times, and lower computational complexities make them a desirable choice for TSC. Although DNN methods have progressed quickly for TSC applications, they still lack generalizability compared to traditional methods. 
Still, based on the relative complexity of the DNN-based methods, we can divide them into two subcategories Simple (earlier, low computational cost, less accurate methods) or Complex (latest, higher complexity, more accurate methods).

\subsubsection{\textbf{Simple DNN-based TSC}}
Early DNN-based TSC approaches began with the simplest method, MultiMayer Perceptron (MLP). MLP is composed of four Fully Connected (FC) layers and was proposed as a baseline for TSC. Multi-scale Convolutional Neural Network (MCNN) was composed of two convolutional layers with max pooling and two FC layers~\cite{MCNN}. Even though MCNN was simple, it required heavy and complex data preprocessing steps. Time-LENET~\cite{t-LENET} had similar architecture to MCNN but with a modified pooling method. Time-CNN~\cite{time-cnn} used Mean Squared Error (MSE) loss for training its model and removed the final Global Average Pooling (GAP) layer behind the FC layer. Multi-Channel Deep Convolutional Neural Network (MCDCNN)~\cite{MCDCNN} also had a similar architecture designed for multivariate data. It applied parallel and independent convolutions to each input channel. 
Recurrent DNNs were traditionally used for time series forecasting in the form of Echo State Networks (ESNs). Time Warping Invariant Echo State Network (TWIESN)~\cite{ TWIESN} was a recurrent DNN method that redesigned ESNs for TSC.

\subsubsection{\textbf{Complex DNN-based TSC}}
The performance of early DNN-based TSC methods, based on accuracy benchmarking on UCR/UEA datasets, was still inferior to traditional methods~\cite{rank1, 2021hive}. The introduction of more complex 1D-CNN architectures (such as FCN and ResNet) showed that new DNN methods could achieve similar results to traditional methods with lower computational complexities and training times. FCN model was a three-layered 1D-CNN model that preserved the length of the series throughout all its layers. The output layer was a fully connected layer right after global average pooling (GAP). The GAP layer was later replaced with an Attention layer in~\cite{encoder}. Residual Network (ResNet) is a deeper model with eleven 1D convolutional layers. ResNet was constructed with three residual blocks followed by GAP and a softmax classifier. The inferior performances of FCN and ResNet on UCR datasets compared to HIVE-COTE~\cite{rank1,dempster2020rocket} meant that DNN-based TSCs could still be improved.

Inception Time~\cite{fawaz2020inceptiontime} was introduced as a DNN-based TSC method that achieved comparable accuracy to HIVE-COTE. It was developed as a TSC equivalent of the image classification architecture, AlexNet. It consisted of multiple inception modules~\cite{szegedy2015going} that apply four concurrent convolutional filters of varying kernel sizes on their input. OS-CNN~\cite{tang_omni-scale_2022} is a newer 1D-CNN deep learning-based TSC method. It used OS-Blocks in which the kernel size of each layer is different based on the data. OS-CNN’s results for UCR benchmark were marginally better than InceptionTime. However, its performance on UEA benchmark was worse than FCN, ResNet, and InceptionTime. DA-Net~\cite{chen_da-net_2022} is a model composed of two layers of Squeeze Excitation Window Attention (SEWA) and the Sparse SelfAttention within Windows (SSAW). The first layer is a 1D version of squeeze and excitation (SE) block~\cite{hu2018squeeze}, and SSAW is a windowed multi-head attention~\cite{liu2021swin} layer. Even though this model utilized both self-attention and temporal attention, its results on the UEA benchmark are significantly worse than OS-CNN and, subsequently, FCN, ResNet, and InceptionTime. Voice2Series~\cite {yang_voice2series_2021} leverages large-scale pre-trained speech models by reprogramming the input time series. Its performance was only reported on 30 out of 128 datasets of the UCR benchmark. Therefore, the generality of this method is somewhat questionable.

A few works focus on changing the convolution-based methods to a more suitable approach for TSC. DTWNet~\cite{cai_dtwnet_2019} replaces the inner product kernel with a DTW kernel. However, the authors evaluated its performance entirely differently from other related works.~\cite{ouyang_convolutional_2021} replaced the inner dot product between the kernels and the input by Elastic Matching (EM) Mechanism in the form of an FC layer that imitates DTW. Therefore, the model became invariant to time distortion.\\

TapNet~\cite{tapnet}, SimTSC~\cite{noauthor_towards_nodate}, SelfMatch~\cite{xing_selfmatch_nodate} and iTimes~\cite{liu_itimes_2022} are a few works from the scope of semi-supervised TSC with different testing methods. These methods combine traditional TSC and feature prototyping to utilize unlabeled data. Since these methods require extensive external data for training, their results on isolated UEA/UCR benchmarks were lower than supervised methods.\\

FCN, ResNet, and InceptionTime have been demonstrated to be the most successful methods for TSC by achieving the highest ranks~\cite{rank1, rank2} on the UCR archive~\cite{dau2019ucr}. They were therefore chosen as the baseline models for the work presented here. We would have considered using more models as our base models, such as XCM~\cite{xcm}, ShapeNet (SN)~\cite{shapenet}), and TCRAN~\cite{tcran}. However, their reported performances on the UEA dataset included either marginal or no improvement compared to the FCN’s. This statement concludes the review of TSC work related to our proposed method. However, since the proposed TPS model is a transformer with a positional information injection module, it would be essential to review related works on positional information modification in transformers.

\subsection{Positional information injection in transformers}
Our TPS model is a transformer with a modified approach to processing positional information. Therefore, This section reviews the related works on positional information injection in transformers. Transformer models~\cite{vaswani2017attention} have shown good performance for many natural language processing tasks. The baseline Self-Attention (SA) transformer model is indifferent to the time order of the input. However, text data is inherently sequential. Therefore, the injection of position information in transformers is the focus of many methods.~\cite{dufter_position_2022} provided an overview of these methods. It laid out multiple categorizing specifications such as (1) Reference Point (Ref. P): Absolute (Abs) or Relative (Rel) position information, (2) Injection Method (Inj. M): Additive Positional Embedding (APE) or Manipulating Attention Matrices (MAM), (3) Learnable during training or Fixed. Based on these categories, our proposed TPS algorithm could fall into the Relative, MAM, and Learnable positional information processing method group. In the next paragraph, we provide an overview of the methods in this field. The novelty of these methods is in positional information injection into transformers.

\cite{shaw_self-attention_2018} modified a self-attention matrix by adding a learned representation of relative positions using the distance between time entries.~\cite{shaw_self-attention_2018} hypothesized that the exact relative positional information is not useful beyond a certain distance.
DeBERTa~\cite{he2021deberta} represented each word by two vectors of content and position. The positional vectors were used to generate a second attention matrix added to the original.~\cite{he2021deberta} also injected a traditional absolute position embedding into its last stage, utilizing APE and MAM injection methods and Abs and Rel position to embedding conversion. TUPE~\cite{ke_rethinking_2021} separated the analysis of position and content. Both relative and absolute positional placements were used to create a position-based attention matrix, added to the separately calculated content correlation attention matrix. SPE~\cite{pmlr-v139-liutkus21a} proposed a combination of $K$ learned sinusoidal components to replace classical additive fixed Positional Embedding in sparse transformers.
\begin{table}[!t]
\setlength{\tabcolsep}{4pt}
  \centering
  \caption{Comparison between the number of parameters of positional information processing methods.}
\begin{tabular}{l|ccc}
\toprule
Method & Ref.P & Inj.M & No. Parameters \\
\midrule
Shaw et al~\cite{shaw_self-attention_2018} & Rel & MAM & \(2(2N − 1)dl\)  \\
Huang et al~\cite{huang-etal-2020-improve} & Rel & MAM & \(dlh(2N − 1)\)  \\
DeBERTa~\cite{he2021deberta} & Rel+Abs & MAM+APE & \(3Nd\)  \\
Transformer XL~\cite{dai-etal-2019-transformer} & Rel & MAM & \(2d + d^2 lh\)  \\
DA-transformer~\cite{wu_da-transformer_2021} & Rel & MAM & \(2dlh\)  \\
TUPE~\cite{ke_rethinking_2021} & Rel+Abs & MAM & \(2h\)  \\
SPE~\cite{pmlr-v139-liutkus21a} & Rel & MAM & \(3Kdh + dl\)  \\
TPS(ours) & Rel & MAM & \(2(d^2/h + d)l\)  \\
TPS+PE (ours) & Rel+Abs & MAM+APE & \(2d^2l/h + (2N+2l)d\)  \\
\bottomrule
\end{tabular}%

  \label{tb:nop}
 \end{table}
 
\cite{wu_da-transformer_2021} proposed a direct relative and multiplicative smoothing on the attention matrix.~\cite{huang-etal-2020-improve} took on a similar approach. But it included both Rel and Abs reference point utilization. A summary of the number of Ref. P, Inj. M, and learnable parameters for these methods is presented in Table~\ref{tb:nop}. In this table, $N$ refers to the max sequence length, $h$ is the number of attention heads, $l$ represents the number of layers, and $d$ is the input dimension size. 

It is very hard to quantitively compare these methods as they seemed to be used for different tasks and tested using different datasets. However, none of these methods is used in applications of TSC tasks. 

\section{Approach}
This section describes and details each proposed attention block's operation format.
\subsection{Global Temporal Attention block (GTA)}
Temporal Attention (TA) is useful for TSC and regression-related problems~\cite{doughty2019pros, zeng2020hybrid, xu2019learning}. In a Classic TA (CTA) block, features from each time unit emphasize or suppress the content based on how informative they are in creating class separation. However, CTA block has two limitations. 

First, the attention weight calculation for each time sample is dependent only on the values at that time, as shown in Eq.~\eqref{eq:1}.
\begin{equation} \label{eq:1}
A =\sigma_1\ (W_2\sigma_2\ (W_1F^T)).
\end{equation}
In this equation, the input time series array is $F=\left(f_1,f_2,\ldots,f_N\right)$ with a dimension of $d\times N$. $d$ refers to the input's dimension size and $N$ refers to input's max length (duration). Therefore, the dimensions for weights $W_1$ and $W_2$ are $\frac{d}{r}\times d$ and $1\times \frac{d}{r}$, respectively. $W_1$ is dimensionality-reduction layer, in which the input dimension $d$ is decreased by a factor of $r$ (dimensionality-reduction factor).
$ \sigma_1\left(.\right)$ and $ \sigma_2\left(.\right)$ indicate \textit{softmax} and \textit{ReLU} activation functions. The attention matrix $A \in \mathbb{R}^{1 \times N}$ and the output of the attention block, $O\in \mathbb{R}^{d \times N}$, is shown in Eq.~\eqref{eq:2}.\\
\begin{equation} \label{eq:2}
O =\left(o_1, o_2 ,\ldots, o_N\right)=F\times diag\left(A\right).
\end{equation}
As shown, the temporal relations between time samples do not affect the calculated attention weights (each $o_i$ is multiplied by a number between 0 and 1, which is only dependent on $f_i$).

Second, each time sample's temporal location has no impact on the calculated attention weights. Some temporal samples may be more important than others due to their temporal location. CTA block does not seem to factor in such importance when calculating attention weights. 

We propose a novel Global Temporal Attention (GTA) block to address these two limitations. The formula for calculating the global attention is shown in Eq.~\eqref{eq:3}. In this equation, learnable weights $W_1$, $W_2$, and $W_3$ have dimensions of $1\times d$, $ \frac{T}{r}\times T$, and $T\times\frac{T}{r}$. $W_2$ and $W_3$ apply a dimensionality-decrease /increase with the reduction/increase coefficient of $r$ set to a default value of 16. $ \sigma_1\left(.\right)$ and $ \sigma_2\left(.\right)$ depict sigmoid and ReLU activations. $A_1$ has a dimension of $1\ \times T$. The final output of GTA block, $ O $, is calculated in the same manner as shown in Eq.~\eqref{eq:2}.
\begin{equation} \label{eq:3}
\begin{split}
A_1 &=\sigma_2\left(W_1F^T\right),\: and\\
A^T &=\sigma_1\left(W_3\sigma_2\left(W_2A_1^T\right)\right).
\end{split}
\end{equation}

\begin{figure}[!t]
\centering
\includegraphics[width=0.40\textwidth]{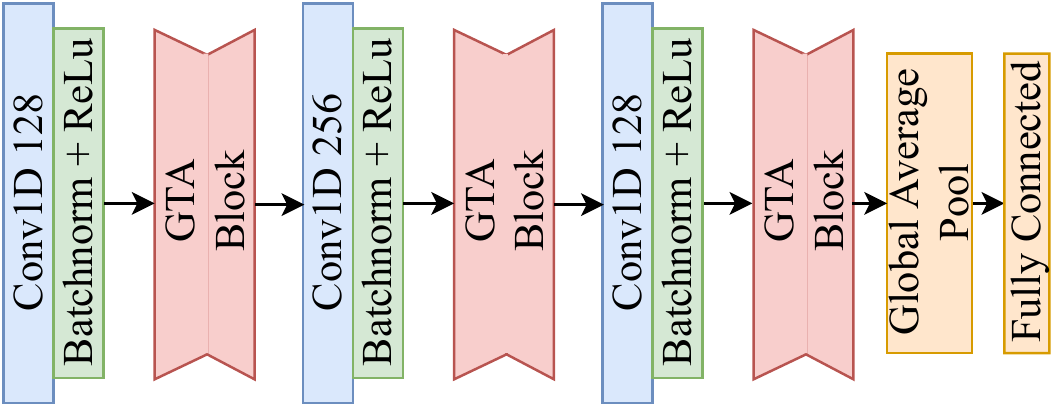}
\caption{FCN augmented with GTA blocks after each processing layer. }
\label{fig:fcngta}
\end{figure}

A GTA block learns to utilize global temporal information to emphasize informative time samples and suppress non-informative ones. Its structure enables determining samples' attention based on their temporal location instead of their values exclusively. Therefore, temporal relations between time samples and their placements are used to determine the importance of time samples during the model's training. Given the similarities between CTA and GTA blocks with the squeeze and excitation (SE) block~\cite{hu2018squeeze}, a GTA block was added as an intermediate block after each processing layer. An example of such use for FCN model~\cite{wang2017time} is shown in Fig.~\ref{fig:fcngta}. One potential limitation of GTA could be its susceptibility to misclassification from temporal shifts. This is related to how GTA processes temporal placement information. During training, $W_1$ in Eq.~\eqref{eq:3} is fixated on values that depend on temporal placements of the training data. Therefore, a system that can overcome such potential limitation is needed.
\subsection{Temporal Pseudo-Gaussian augmented Self-attention}
The self-attention mechanism successfully replaced recurrence in the field of language modeling~\cite{vaswani2017attention}. The similarities between the two fields of language modeling and time series analysis suggest that self-attention might be a promising method for TSC. self-attention in MTSC has been explored by~\cite{liu2021gated, zerveas2020transformer}. However, two main reasons motivated a reformulation of the attention calculation in the self-attention mechanism for TSC. 

\begin{figure}[!t]
\centering
\includegraphics[width=0.40\textwidth  ]{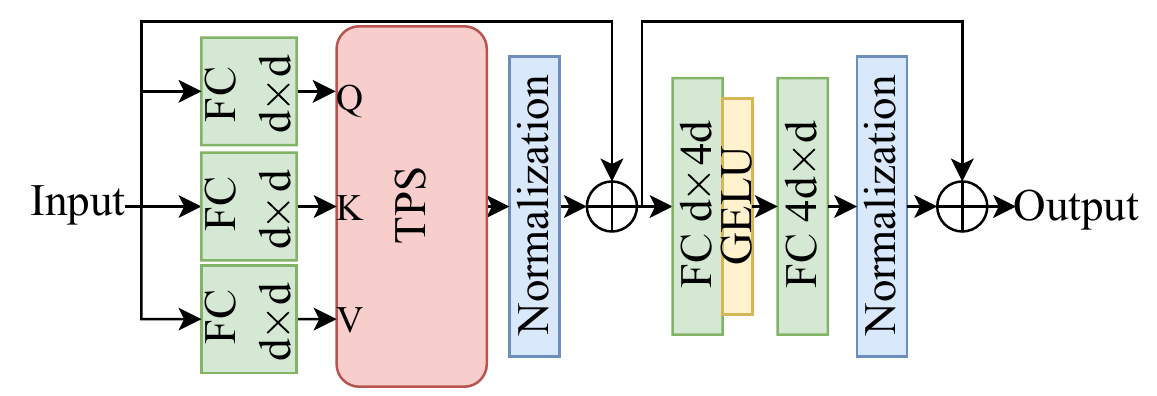}
\caption{The complete TPS encoder model's block diagram.}
\label{fig:enc}
\end{figure}

In the reformulated method, the calculation of attention weights for each time sample is not limited only to the relative similarity of that sample's content with the other samples; it is also dependent on its relative positional placement. The attention weights are then modified so that more consideration will be given to the neighboring samples based on the content of the current sample in a pseudo-Gaussian distribution form. We call this distribution pseudo-Gaussian because it is similar to Gaussian, but it is not symmetric. Moreover, its distribution is normalized after combining it with the self-attention weights.
\begin{equation} \label{eq:kvq}
\begin{split}
F^T =(f_1,f_2,& f_3,\ldots,f_N ),\\
K^T= W_K F^T,\: Q^T=& W_Q F^T,\: V^T= W_V F^T,\\
W_K,\ W_Q,&\ W_V \in \mathbb{R}^{d \times d}.
\end{split}
\end{equation}
The proper inputs for the self-attention mechanism are generated by transforming the input time series ($F\in \mathbb{R}^{N \times d}$) into three elements of $Q$, $K$, and $V$ as described by Eq~\eqref{eq:kvq}. $Q$, $K$, and $V$ stand for query, key, and value. They each have a $N\times d$ dimension, where $N$ indicates the Maximum sequence length and $d$ is the feature array length. This operation is shown in Fig~\ref{fig:enc} as passing the input through three fully connected layers. Then, The self-attention mechanism transforms the query and the set of key-value pairs into an output, as described in Eq.~\eqref{eq:sa}. The output of self-attention $O$ is calculated by multiplying $V$ by $A \in (\mathbb{R}^{N \times N})$.

\begin{equation} \label{eq:sa}
\begin{split}
K^T &=\left(k_1,k_2,k_3,\ldots,k_N\right),\\
Q^T &=\left(q_1,q_2,q_3,\ldots,q_N\right),\\
V^T &=\left(v_1,v_2,v_3,\ldots,v_N\right),\\
A &=Softmax\left(\frac{QK^T}{\sqrt d}\right),\: and\\
O&=AV.
\end{split}
\end{equation}

The formulation for TPS is shown in Eq.~\eqref{eq:tps} in which the attention matrix calculation is modified. First, a scaling function is applied to the base attention matrix ($S\left(.\right)$). Second, the scaled attention matrix ($A_1$) is combined with the new pseudo-Gaussian temporal attention matrix ($A_2$). Finally, the result of this addition is normalized ($\mathbb{N}\left(.\right)$) by dividing each row by the sum of its elements.
\begin{equation} \label{eq:tps}
\begin{split}
A_1&=S\left(Softmax\left(\frac{QK^T}{\sqrt d}\right)\right),\\
A_2^T&=\left(P_1,P_2,\ldots,P_N\right),\\
A&=\mathbb{N}\left(\frac{A_1+ A_2}{2}\right), and\\
O&=AV.
\end{split}
\end{equation}

The calculation for the additional pseudo-Gaussian temporal attention matrix $A_2$ is presented in Eq.~\eqref{eq:tps2}. Each row of $A_2$ is presented by $P_i$, where $i$ indicates the row number. $p_{i,j}$ represents the element that modifies the attention weight between time samples $i$ and $j$ based on their distance. However, this relation does follow a similar pseudo-gaussian distribution. The Gaussian variance would be different if time sample $j$ is placed before or after time-sample $i$, $\hat{\sigma}_i^{2}$, and $\sigma_i^2$, respectively. Additionally, both $\hat{\sigma}_i$, and $\sigma_i$ are calculated based on $v_i$, the value of time sample $i$. In Eq~\eqref{eq:tps2} $W'$ and $W$ are $1\times d$ dimensional learnable weight matrices, and $b$ is a configurable bias determined empirically. 
\begin{equation} \label{eq:tps2}
\begin{split}
\hat{\sigma_i}&=\left|W^\prime v_i\right|+b, \\
\sigma_i&=\left|Wv_i\right|+b,\\
p_{i,j}&=\begin{dcases}
e^{-\dfrac{1}{2}\dfrac{i-j}{2\hat{\sigma}^{2}_{i}} }, &j <i\\
e^{-\dfrac{1}{2}\dfrac{i-j}{2\sigma^{2}_{i}} }, &j\geq i\\
\end{dcases},\\
P_i^T&=\left(p_{i,1},\ p_{i,2},p_{i,3},\ ...,\ p_{i,T}\right),\: and\\
A_2^T&=\left(P_1,P_2,\ldots,P_T\right).\\
\end{split}
\end{equation}

\begin{figure}[!t]
\centering
\includegraphics[width=0.45\textwidth]{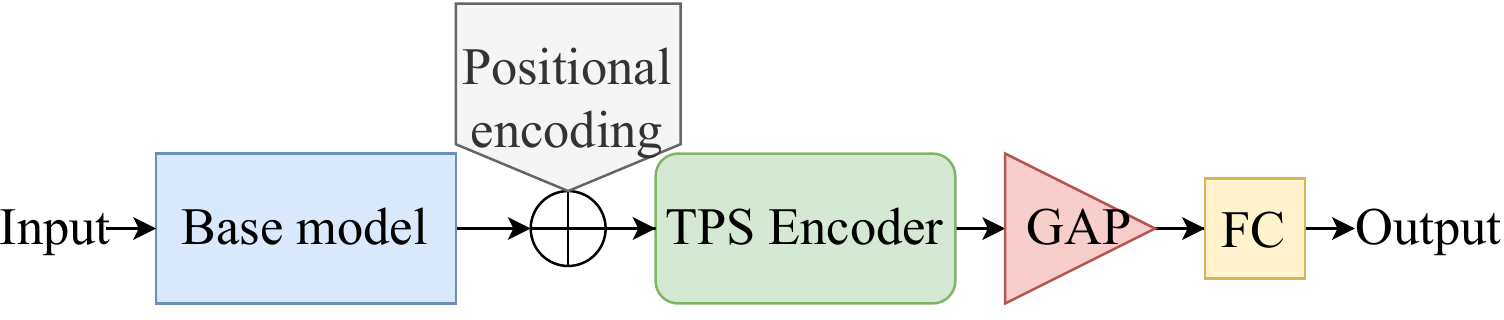}
\caption{Typical usage of the TPS encoder.}
\label{fig:st}
\end{figure}

The complete TPS processing structure is shown in Fig.~\ref{fig:enc}. It is inspired by an encoder structure first presented in~\cite{vaswani2017attention}. 
The suggested application for TPS for incorporating it into a base TSC model is shown in Fig.~\ref{fig:st}. It is independent of the TSC model's architecture. TPS can be integrated into a model as simple as a single FC layer or as complicated as an InceptionTime~\cite{fawaz2020inceptiontime}. Our method enables general users to enhance the performance of an existing model by simply adding the TPS module. Positional encoding (PE) allows the model to utilize sequence order by adding information about the Absolute position of each time sample into its embedding. We used learnable positional encoding introduced in~\cite{gehring2017convolutional} to project positional information into the input array (shown by $\oplus$ in Fig.~\ref{fig:st}). PE injection is placed after the base model, similar to its placement (after the embedding layer/FC layer) in~\cite{ vaswani2017attention}. Unlike self-attention encoders, 1D-CNN TSC classifiers do not need positional encoding. As the convolutional and kernel operation inherently process the positional placements into the outcome. So, PE is only required to be injected before the data is entered into the self-attention layer.

\section{Experimental Results}
We explored different experimental settings to evaluate and compare GTA and TPS on different applications. 

\begin{table*}[!t]
\setlength{\tabcolsep}{2pt}
  \centering
  \caption{Accuracy [\%] comparison between state-of-the-art baseline TSC models and proposed GTA, TPS, and PE blocks on UEA benchmark datasets.}

\begin{tabular}{l|c|c|c|c|c|cccc|cccc|cccc|}
\toprule
Base  & DA-Net & Tap-Net & OC-CNN & XCM   & SN    & \multicolumn{4}{c|}{FCN}      & \multicolumn{4}{c|}{RESNET}   & \multicolumn{4}{c|}{InceptionTime} \\
      & \cite{chen_da-net_2022}  & \cite{tapnet} & \cite{tang_omni-scale_2022} & \cite{xcm} & \cite{shapenet} & \multicolumn{4}{c|}{}         & \multicolumn{4}{c|}{}         & \multicolumn{4}{c|}{} \\
\midrule
GTA   & \xmark & \xmark & \xmark & \xmark & \xmark & \multicolumn{1}{c|}{\xmark} & \multicolumn{1}{c|}{\cmark} & \multicolumn{1}{c|}{\xmark} & \xmark & \multicolumn{1}{c|}{\xmark} & \multicolumn{1}{c|}{\cmark} & \multicolumn{1}{c|}{\xmark} & \xmark & \multicolumn{1}{c|}{\xmark} & \multicolumn{1}{c|}{\cmark} & \multicolumn{1}{c|}{\xmark} & \xmark \\
TPS   & \xmark & \xmark & \xmark & \xmark & \xmark & \multicolumn{1}{c|}{\xmark} & \multicolumn{1}{c|}{\xmark} & \multicolumn{1}{c|}{\cmark} & \cmark & \multicolumn{1}{c|}{\xmark} & \multicolumn{1}{c|}{\xmark} & \multicolumn{1}{c|}{\cmark} & \cmark & \multicolumn{1}{c|}{\xmark} & \multicolumn{1}{c|}{\xmark} & \multicolumn{1}{c|}{\cmark} & \cmark \\
PE    & \xmark & \xmark & \xmark & \xmark & \xmark & \multicolumn{1}{c|}{\xmark} & \multicolumn{1}{c|}{\xmark} & \multicolumn{1}{c|}{\cmark} & \xmark & \multicolumn{1}{c|}{\xmark} & \multicolumn{1}{c|}{\xmark} & \multicolumn{1}{c|}{\cmark} & \xmark & \multicolumn{1}{c|}{\xmark} & \multicolumn{1}{c|}{\xmark} & \multicolumn{1}{c|}{\cmark} & \xmark \\
\midrule
ArticularyWordRecognition (AWR) & 98    & 98.7  & 98.8  & 98.3  & 98.7  & \textcolor[rgb]{ 1,  0,  0}{98.3} & \textcolor[rgb]{ .188,  .329,  .588}{98.7} & 98.0  & \textcolor[rgb]{ .188,  .329,  .588}{98.7} & \textcolor[rgb]{ 1,  0,  0}{98.3} & \textcolor[rgb]{ .776,  .349,  .067}{\textbf{99.0}} & 98.3  & 98.3  & \textcolor[rgb]{ 1,  0,  0}{98.5} & \textcolor[rgb]{ .439,  .678,  .278}{\textbf{99.0}} & \textcolor[rgb]{ .439,  .678,  .278}{\textbf{99.0}} & \textcolor[rgb]{ .439,  .678,  .278}{\textbf{99.0}} \\
AtrialFibrillation (AF) & 46.7  & 33.3  & 23.3  & 46.7  & 40    & \textcolor[rgb]{ 1,  0,  0}{40.0} & 40.0  & 40.0  & \textcolor[rgb]{ .188,  .329,  .588}{46.7} & \textcolor[rgb]{ 1,  0,  0}{40.0} & 33.3  & 40.0  & \textcolor[rgb]{ .776,  .349,  .067}{46.7} & \textcolor[rgb]{ 1,  0,  0}{40.0} & \textcolor[rgb]{ .439,  .678,  .278}{46.7} & 40.0  & \textcolor[rgb]{ .439,  .678,  .278}{\textbf{53.3}} \\
BasicMotions (BM) & 92.5  & \textbf{100} & \textbf{100.0} & \textbf{100} & \textbf{100} & \textcolor[rgb]{ 1,  0,  0}{\textbf{100.0}} & \textbf{100.0} & \textbf{100.0} & \textbf{100.0} & \textcolor[rgb]{ 1,  0,  0}{\textbf{100.0}} & \textbf{100.0} & \textbf{100.0} & \textbf{100.0} & \textcolor[rgb]{ 1,  0,  0}{\textbf{100.0}} & \textbf{100.0} & \textbf{100.0} & \textbf{100.0} \\
CharacterTrajectories (CT) & 99.8  & 99.7  & 99.8  & 99.5  & 98    & \textcolor[rgb]{ 1,  0,  0}{99.3} & \textcolor[rgb]{ .188,  .329,  .588}{99.5} & 99.1  & \textcolor[rgb]{ .188,  .329,  .588}{99.6} & \textcolor[rgb]{ 1,  0,  0}{99.3} & \textcolor[rgb]{ .776,  .349,  .067}{99.5} & 99.3  & \textcolor[rgb]{ .776,  .349,  .067}{99.5} & \textcolor[rgb]{ 1,  0,  0}{99.9} & \textcolor[rgb]{ .439,  .678,  .278}{\textbf{100.0}} & \textcolor[rgb]{ .439,  .678,  .278}{99.9} & 99.8 \\
Cricket (CR) & 86.1  & 95.8  & 99.3  & \textbf{100} & 98.6  & \textcolor[rgb]{ 1,  0,  0}{98.6} & \textcolor[rgb]{ .188,  .329,  .588}{\textbf{100.0}} & \textcolor[rgb]{ .188,  .329,  .588}{98.6} & \textcolor[rgb]{ .188,  .329,  .588}{\textbf{100.0}} & \textcolor[rgb]{ 1,  0,  0}{98.6} & \textcolor[rgb]{ .776,  .349,  .067}{98.6} & \textcolor[rgb]{ .776,  .349,  .067}{98.6} & \textcolor[rgb]{ .776,  .349,  .067}{98.6} & \textcolor[rgb]{ 1,  0,  0}{\textbf{100.0}} & \textbf{100.0} & 98.6  & \textbf{100.0} \\
DuckDuckGeese (DDG) & 52.1  & 57.5  & 54.0  & 70    & 72.5  & \textcolor[rgb]{ 1,  0,  0}{69.0} & \textcolor[rgb]{ .188,  .329,  .588}{74.0} & \textcolor[rgb]{ .188,  .329,  .588}{\textbf{78.0}} & \textcolor[rgb]{ .188,  .329,  .588}{76.0} & \textcolor[rgb]{ 1,  0,  0}{67.0} & \textcolor[rgb]{ .776,  .349,  .067}{68.0} & \textcolor[rgb]{ .776,  .349,  .067}{68.0} & \textcolor[rgb]{ .776,  .349,  .067}{72.0} & \textcolor[rgb]{ 1,  0,  0}{66.7} & \textcolor[rgb]{ .439,  .678,  .278}{70.0} & 66.0  & 66.0 \\
EigenWorms (EW) & 48.9  & 48.9  & 41.4  & 43.5  & 87.8  & \textcolor[rgb]{ 1,  0,  0}{54.7} & 51.1  & \textcolor[rgb]{ .188,  .329,  .588}{60.3} & \textcolor[rgb]{ .188,  .329,  .588}{60.3} & \textcolor[rgb]{ 1,  0,  0}{50.6} & \textcolor[rgb]{ .776,  .349,  .067}{51.9} & 50.4  & \textcolor[rgb]{ .776,  .349,  .067}{51.9} & \textcolor[rgb]{ 1,  0,  0}{76.3} & 65.6  & \textcolor[rgb]{ .439,  .678,  .278}{\textbf{94.7}} & \textcolor[rgb]{ .439,  .678,  .278}{93.1} \\
Epilepsy (EP) & 83.3  & 97.1  & 98.0  & \textbf{99.3} & 98.7  & \textcolor[rgb]{ 1,  0,  0}{96.7} & \textcolor[rgb]{ .188,  .329,  .588}{97.1} & \textcolor[rgb]{ .188,  .329,  .588}{98.6} & \textcolor[rgb]{ .188,  .329,  .588}{98.6} & \textcolor[rgb]{ 1,  0,  0}{97.1} & \textcolor[rgb]{ .776,  .349,  .067}{98.6} & \textcolor[rgb]{ .776,  .349,  .067}{97.8} & \textcolor[rgb]{ .776,  .349,  .067}{98.6} & \textcolor[rgb]{ 1,  0,  0}{97.3} & 97.1  & \textcolor[rgb]{ .439,  .678,  .278}{98.6} & 97.1 \\
ERing (EC) & 87.4  & 13.3  & 88.1  & 13.3  & 13.3  & \textcolor[rgb]{ 1,  0,  0}{89.8} & 87.0  & 83.7  & 86.3  & \textcolor[rgb]{ 1,  0,  0}{83.5} & \textcolor[rgb]{ .776,  .349,  .067}{87.8} & 82.2  & 83.0  & \textcolor[rgb]{ 1,  0,  0}{91.7} & \textcolor[rgb]{ .439,  .678,  .278}{93.3} & \textcolor[rgb]{ .439,  .678,  .278}{95.6} & \textcolor[rgb]{ .439,  .678,  .278}{\textbf{97.1}} \\
EthanolConcentration (ER) & 33.8  & 32.3  & 24.0  & 34.6  & 31.2  & \textcolor[rgb]{ 1,  0,  0}{32.1} & \textcolor[rgb]{ .188,  .329,  .588}{32.7} & \textcolor[rgb]{ .188,  .329,  .588}{\textbf{35.4}} & \textcolor[rgb]{ .188,  .329,  .588}{35.0} & \textcolor[rgb]{ 1,  0,  0}{30.0} & \textcolor[rgb]{ .776,  .349,  .067}{33.5} & \textcolor[rgb]{ .776,  .349,  .067}{33.8} & \textcolor[rgb]{ .776,  .349,  .067}{32.7} & \textcolor[rgb]{ 1,  0,  0}{32.1} & \textcolor[rgb]{ .439,  .678,  .278}{34.2} & \textcolor[rgb]{ .439,  .678,  .278}{\textbf{35.4}} & \textcolor[rgb]{ .439,  .678,  .278}{33.8} \\
FaceDetection (FD) & 64.8  & 55.6  & 57.5  & 63.9  & 60.2  & \textcolor[rgb]{ 1,  0,  0}{56.5} & \textcolor[rgb]{ .188,  .329,  .588}{59.7} & 56.0  & 55.6  & \textcolor[rgb]{ 1,  0,  0}{54.2} & \textcolor[rgb]{ .776,  .349,  .067}{56.4} & \textcolor[rgb]{ .776,  .349,  .067}{56.3} & \textcolor[rgb]{ .776,  .349,  .067}{55.9} & \textcolor[rgb]{ 1,  0,  0}{63.6} & \textcolor[rgb]{ .439,  .678,  .278}{65.3} & \textcolor[rgb]{ .439,  .678,  .278}{65.2} & \textcolor[rgb]{ .439,  .678,  .278}{\textbf{65.4}} \\
FingerMovements (FM) & 51    & 53    & 56.8  & 60    & 58    & \textcolor[rgb]{ 1,  0,  0}{58.5} & 57.0  & \textcolor[rgb]{ .188,  .329,  .588}{64.0} & \textcolor[rgb]{ .188,  .329,  .588}{62.0} & \textcolor[rgb]{ 1,  0,  0}{57.5} & \textcolor[rgb]{ .776,  .349,  .067}{59.0} & \textcolor[rgb]{ .776,  .349,  .067}{64.0} & 55.0  & \textcolor[rgb]{ 1,  0,  0}{58.0} & \textcolor[rgb]{ .439,  .678,  .278}{64.0} & 58.0  & \textcolor[rgb]{ .439,  .678,  .278}{\textbf{65.4}} \\
HandMovementDirection (HMD) & 36.5  & 37.8  & 44.3  & 44.6  & 33.8  & \textcolor[rgb]{ 1,  0,  0}{40.5} & \textcolor[rgb]{ .188,  .329,  .588}{45.9} & \textcolor[rgb]{ .188,  .329,  .588}{47.3} & \textcolor[rgb]{ .188,  .329,  .588}{43.2} & \textcolor[rgb]{ 1,  0,  0}{41.2} & \textcolor[rgb]{ .776,  .349,  .067}{44.6} & \textcolor[rgb]{ .776,  .349,  .067}{44.6} & \textcolor[rgb]{ .776,  .349,  .067}{43.2} & \textcolor[rgb]{ 1,  0,  0}{\textbf{48.6}} & 45.9  & \textbf{48.6} & 47.3 \\
Handwriting (HW) & 15.9  & 35.7  & \textbf{66.8} & 41.2  & 45.1  & \textcolor[rgb]{ 1,  0,  0}{39.5} & \textcolor[rgb]{ .188,  .329,  .588}{48.2} & \textcolor[rgb]{ .188,  .329,  .588}{41.9} & 39.3  & \textcolor[rgb]{ 1,  0,  0}{53.1} & \textcolor[rgb]{ .776,  .349,  .067}{53.8} & 44.2  & 43.4  & \textcolor[rgb]{ 1,  0,  0}{57.1} & \textcolor[rgb]{ .439,  .678,  .278}{58.7} & 55.8  & 56.7 \\
Heartbeat (HB) & 62.4  & 75.1  & 48.9  & 77.6  & 75.6  & \textcolor[rgb]{ 1,  0,  0}{77.8} & \textcolor[rgb]{ .188,  .329,  .588}{\textbf{80.5}} & \textcolor[rgb]{ .188,  .329,  .588}{78.5} & \textcolor[rgb]{ .188,  .329,  .588}{78.5} & \textcolor[rgb]{ 1,  0,  0}{75.9} & \textcolor[rgb]{ .776,  .349,  .067}{80.0} & \textcolor[rgb]{ .776,  .349,  .067}{76.6} & \textcolor[rgb]{ .776,  .349,  .067}{77.1} & \textcolor[rgb]{ 1,  0,  0}{77.1} & \textcolor[rgb]{ .439,  .678,  .278}{78.0} & \textcolor[rgb]{ .439,  .678,  .278}{77.6} & \textcolor[rgb]{ .439,  .678,  .278}{78.5} \\
InsectWingbeat (IW) & 56.7  & 20.8  & 66.7  & 10.5  & 25    & \textcolor[rgb]{ 1,  0,  0}{66.0} & \textcolor[rgb]{ .188,  .329,  .588}{66.5} & 63.6  & 62.8  & \textcolor[rgb]{ 1,  0,  0}{62.2} & \textcolor[rgb]{ .776,  .349,  .067}{62.4} & 60.0  & 60.3  & \textcolor[rgb]{ 1,  0,  0}{69.3} & \textcolor[rgb]{ .439,  .678,  .278}{\textbf{69.4}} & \textcolor[rgb]{ .439,  .678,  .278}{69.4} & 69.1 \\
JapaneseVowels (JV) & 93.8  & 96.5  & 99.1  & 98.6  & 98.4  & \textcolor[rgb]{ 1,  0,  0}{98.9} & \textcolor[rgb]{ .188,  .329,  .588}{\textbf{99.7}} & \textcolor[rgb]{ .188,  .329,  .588}{98.9} & \textcolor[rgb]{ .188,  .329,  .588}{98.9} & \textcolor[rgb]{ 1,  0,  0}{98.2} & \textcolor[rgb]{ .776,  .349,  .067}{99.5} & \textcolor[rgb]{ .776,  .349,  .067}{98.9} & \textcolor[rgb]{ .776,  .349,  .067}{99.2} & \textcolor[rgb]{ 1,  0,  0}{98.7} & \textcolor[rgb]{ .439,  .678,  .278}{99.2} & 98.6  & 98.6 \\
Libras (LIB) & 80    & 85    & \textbf{95.0} & 84.4  & 85.6  & \textcolor[rgb]{ 1,  0,  0}{81.7} & \textcolor[rgb]{ .188,  .329,  .588}{95.0} & 78.9  & \textcolor[rgb]{ .188,  .329,  .588}{82.2} & \textcolor[rgb]{ 1,  0,  0}{83.1} & \textcolor[rgb]{ .776,  .349,  .067}{83.9} & \textcolor[rgb]{ .776,  .349,  .067}{88.9} & \textcolor[rgb]{ .776,  .349,  .067}{89.4} & \textcolor[rgb]{ 1,  0,  0}{88.3} & \textcolor[rgb]{ .439,  .678,  .278}{90.6} & \textcolor[rgb]{ .439,  .678,  .278}{88.9} & \textcolor[rgb]{ .439,  .678,  .278}{89.4} \\
LSST (LSST) & 56    & 56.8  & 41.3  & 61.2  & 59    & \textcolor[rgb]{ 1,  0,  0}{51.1} & 47.7  & \textcolor[rgb]{ .188,  .329,  .588}{66.4} & \textcolor[rgb]{ .188,  .329,  .588}{67.3} & \textcolor[rgb]{ 1,  0,  0}{68.5} & \textcolor[rgb]{ .776,  .349,  .067}{\textbf{69.1}} & 65.2  & 67.9  & \textcolor[rgb]{ 1,  0,  0}{55.9} & \textcolor[rgb]{ .439,  .678,  .278}{56.9} & \textcolor[rgb]{ .439,  .678,  .278}{62.5} & \textcolor[rgb]{ .439,  .678,  .278}{66.5} \\
MotorImagery (MI) & 50    & 59    & 53.5  & 54    & 61    & \textcolor[rgb]{ 1,  0,  0}{57.0} & \textcolor[rgb]{ .188,  .329,  .588}{\textbf{66.0}} & \textcolor[rgb]{ .188,  .329,  .588}{63.0} & 55.0  & \textcolor[rgb]{ 1,  0,  0}{58.0} & \textcolor[rgb]{ .776,  .349,  .067}{\textbf{66.0}} & \textcolor[rgb]{ .776,  .349,  .067}{63.0} & \textcolor[rgb]{ .776,  .349,  .067}{64.0} & \textcolor[rgb]{ 1,  0,  0}{64.3} & 64.0  & 56.0  & \textcolor[rgb]{ .439,  .678,  .278}{65.8} \\
NATOPS (NA) & 87.8  & 93.9  & 96.8  & 97.8  & 88.3  & \textcolor[rgb]{ 1,  0,  0}{97.5} & \textcolor[rgb]{ .188,  .329,  .588}{98.9} & \textcolor[rgb]{ .188,  .329,  .588}{\textbf{99.4}} & \textcolor[rgb]{ .188,  .329,  .588}{98.9} & \textcolor[rgb]{ 1,  0,  0}{96.9} & \textcolor[rgb]{ .776,  .349,  .067}{97.8} & 95.6  & \textcolor[rgb]{ .776,  .349,  .067}{97.8} & \textcolor[rgb]{ 1,  0,  0}{95.9} & \textcolor[rgb]{ .439,  .678,  .278}{96.7} & \textcolor[rgb]{ .439,  .678,  .278}{96.1} & 94.4 \\
PEMS-SF (PEMS) & 86.7  & 75.1  & 76.0  & 99.1  & 75.1  & \textcolor[rgb]{ 1,  0,  0}{99.0} & 78.6  & \textcolor[rgb]{ .188,  .329,  .588}{\textbf{99.2}} & 79.2  & \textcolor[rgb]{ 1,  0,  0}{98.9} & 77.5  & \textcolor[rgb]{ .776,  .349,  .067}{99.1} & 76.3  & \textcolor[rgb]{ 1,  0,  0}{80.0} & \textcolor[rgb]{ .439,  .678,  .278}{81.5} & \textcolor[rgb]{ .439,  .678,  .278}{81.5} & 78.0 \\
PenDigits (PD) & 98    & 98    & 98.5  & 75.7  & 97.7  & \textcolor[rgb]{ 1,  0,  0}{62.1} & \textcolor[rgb]{ .188,  .329,  .588}{\textbf{99.2}} & \textcolor[rgb]{ .188,  .329,  .588}{78.6} & \textcolor[rgb]{ .188,  .329,  .588}{99.1} & \textcolor[rgb]{ 1,  0,  0}{41.3} & \textcolor[rgb]{ .776,  .349,  .067}{99.0} & \textcolor[rgb]{ .776,  .349,  .067}{80.3} & \textcolor[rgb]{ .776,  .349,  .067}{99.1} & \textcolor[rgb]{ 1,  0,  0}{99.0} & \textcolor[rgb]{ .439,  .678,  .278}{99.1} & 98.9  & \textcolor[rgb]{ .439,  .678,  .278}{99.1} \\
Phoneme (PM) & 9.3   & 17.5  & 29.9  & 22.5  & 29.8  & \textcolor[rgb]{ 1,  0,  0}{28.6} & \textcolor[rgb]{ .188,  .329,  .588}{29.2} & \textcolor[rgb]{ .188,  .329,  .588}{30.7} & \textcolor[rgb]{ .188,  .329,  .588}{30.5} & \textcolor[rgb]{ 1,  0,  0}{33.7} & 33.6  & 31.7  & 30.6  & \textcolor[rgb]{ 1,  0,  0}{\textbf{34.1}} & 34.1  & 33.6  & 32.6 \\
RacketSports (RS) & 80.3  & 86.8  & 87.7  & 89.5  & 88.2  & \textcolor[rgb]{ 1,  0,  0}{88.8} & \textcolor[rgb]{ .188,  .329,  .588}{89.5} & \textcolor[rgb]{ .188,  .329,  .588}{92.8} & \textcolor[rgb]{ .188,  .329,  .588}{90.8} & \textcolor[rgb]{ 1,  0,  0}{87.2} & \textcolor[rgb]{ .776,  .349,  .067}{92.1} & \textcolor[rgb]{ .776,  .349,  .067}{90.8} & \textcolor[rgb]{ .776,  .349,  .067}{\textbf{93.4}} & \textcolor[rgb]{ 1,  0,  0}{90.1} & \textcolor[rgb]{ .439,  .678,  .278}{91.4} & 90.1  & \textcolor[rgb]{ .439,  .678,  .278}{90.8} \\
SelfRegulationSCP1 (SRS1) & 92.4  & 65.2  & 83.5  & 87.8  & 78.2  & \textcolor[rgb]{ 1,  0,  0}{85.5} & \textcolor[rgb]{ .188,  .329,  .588}{\textbf{92.8}} & \textcolor[rgb]{ .188,  .329,  .588}{88.4} & \textcolor[rgb]{ .188,  .329,  .588}{89.1} & \textcolor[rgb]{ 1,  0,  0}{87.4} & \textcolor[rgb]{ .776,  .349,  .067}{90.4} & \textcolor[rgb]{ .776,  .349,  .067}{89.4} & \textcolor[rgb]{ .776,  .349,  .067}{88.7} & \textcolor[rgb]{ 1,  0,  0}{84.6} & \textcolor[rgb]{ .439,  .678,  .278}{87.7} & \textcolor[rgb]{ .439,  .678,  .278}{89.4} & \textcolor[rgb]{ .439,  .678,  .278}{90.1} \\
SelfRegulationSCP2 (SRS2) & 56.1  & 55    & 53.2  & 54.4  & 57.8  & \textcolor[rgb]{ 1,  0,  0}{56.4} & \textcolor[rgb]{ .188,  .329,  .588}{59.4} & \textcolor[rgb]{ .188,  .329,  .588}{60.0} & \textcolor[rgb]{ .188,  .329,  .588}{\textbf{61.1}} & \textcolor[rgb]{ 1,  0,  0}{56.4} & \textcolor[rgb]{ .776,  .349,  .067}{\textbf{61.1}} & \textcolor[rgb]{ .776,  .349,  .067}{60.6} & \textcolor[rgb]{ .776,  .349,  .067}{58.3} & \textcolor[rgb]{ 1,  0,  0}{57.6} & \textcolor[rgb]{ .439,  .678,  .278}{58.9} & 57.2  & \textcolor[rgb]{ .439,  .678,  .278}{59.4} \\
SpokenArabicDigits (SAD) & 95    & 98.3  & 99.7  & 99.5  & 97.5  & \textcolor[rgb]{ 1,  0,  0}{99.3} & \textcolor[rgb]{ .188,  .329,  .588}{99.9} & 99.3  & \textcolor[rgb]{ .188,  .329,  .588}{99.4} & \textcolor[rgb]{ 1,  0,  0}{99.7} & \textcolor[rgb]{ .776,  .349,  .067}{99.8} & 99.2  & \textcolor[rgb]{ .776,  .349,  .067}{99.8} & \textcolor[rgb]{ 1,  0,  0}{99.7} & \textcolor[rgb]{ .439,  .678,  .278}{\textbf{100.0}} & 99.5  & \textcolor[rgb]{ .439,  .678,  .278}{99.8} \\
StandWalkJump (SWJ) & 40    & 40    & 38.3  & 40    & 53.3  & \textcolor[rgb]{ 1,  0,  0}{36.7} & \textcolor[rgb]{ .188,  .329,  .588}{53.3} & \textcolor[rgb]{ .188,  .329,  .588}{\textbf{66.7}} & \textcolor[rgb]{ .188,  .329,  .588}{60.0} & \textcolor[rgb]{ 1,  0,  0}{46.7} & \textcolor[rgb]{ .776,  .349,  .067}{46.7} & \textcolor[rgb]{ .776,  .349,  .067}{46.7} & \textcolor[rgb]{ .776,  .349,  .067}{53.3} & \textcolor[rgb]{ 1,  0,  0}{37.8} & \textcolor[rgb]{ .439,  .678,  .278}{60.0} & \textcolor[rgb]{ .439,  .678,  .278}{46.7} & \textcolor[rgb]{ .439,  .678,  .278}{46.7} \\
UWaveGestureLibrary (UW) & 83.3  & 89.4  & \textbf{92.7} & 89.4  & 90.6  & \textcolor[rgb]{ 1,  0,  0}{79.8} & \textcolor[rgb]{ .188,  .329,  .588}{85.9} & \textcolor[rgb]{ .188,  .329,  .588}{82.5} & 74.7  & \textcolor[rgb]{ 1,  0,  0}{71.6} & \textcolor[rgb]{ .776,  .349,  .067}{79.4} & \textcolor[rgb]{ .776,  .349,  .067}{75.6} & \textcolor[rgb]{ .776,  .349,  .067}{74.1} & \textcolor[rgb]{ 1,  0,  0}{90.5} & \textcolor[rgb]{ .439,  .678,  .278}{90.6} & 89.7  & 90.0 \\
\midrule
Average & 67.5  & 65.7  & 70.4  & 68.6  & 69.9  & \textcolor[rgb]{ 1,  0,  0}{71.3} & \textcolor[rgb]{ .188,  .329,  .588}{74.4} & \textcolor[rgb]{ .188,  .329,  .588}{74.9} & \textcolor[rgb]{ .188,  .329,  .588}{74.3} & \textcolor[rgb]{ 1,  0,  0}{71.2} & \textcolor[rgb]{ .776,  .349,  .067}{74.1} & \textcolor[rgb]{ .776,  .349,  .067}{73.3} & \textcolor[rgb]{ .776,  .349,  .067}{73.6} & \textcolor[rgb]{ 1,  0,  0}{75.1} & \textcolor[rgb]{ .439,  .678,  .278}{76.6} & \textcolor[rgb]{ .439,  .678,  .278}{76.4} & \textcolor[rgb]{ .439,  .678,  .278}{\textbf{77.4}} \\
\midrule
Rank Average & 13.0  & 13.2  & 10.3  & 9.3   & 10.6  & \textcolor[rgb]{ 1,  0,  0}{10.9} & \textcolor[rgb]{ .188,  .329,  .588}{6.4} & 7.1   & 7.0   & \textcolor[rgb]{ 1,  0,  0}{11.0} & \textcolor[rgb]{ .776,  .349,  .067}{6.3} & 8.7   & 7.9   & \textcolor[rgb]{ 1,  0,  0}{7.4} & \textcolor[rgb]{ .573,  .816,  .314}{\textbf{4.2}} & 6.2   & 4.9 \\
\bottomrule
\end{tabular}%

  \label{tb:all}%
\end{table*}%
\subsubsection{Benchmark Datasets}
The UEA Multivariate TSC archive~\cite{bagnall2018uea} is a collection of 30 datasets of various types, dimensions, and lengths series. These datasets are selected from various applications, including human activity recognition and classification of motion, electroencephalogram (ECG)/electroencephalography (EEG)/magnetoencephalogram (MEG) signals, audio spectra, and more. The dimensions of these datasets vary from 2 (AtrialFibrillation, Libras, PenDigits) to 1345 (DuckDuckGeese). The series lengths vary from 8 (PenDigits) to 17984 (EigenWorms) time samples. The collection was introduced as a benchmark for the standardized evaluation of MTSC algorithms. We used the UEA archive to assess and compare the proposed blocks against the state-of-the-art TSC methods. 

\subsubsection{Hyperparameters and Hardware Setup}
Following~\cite{fawaz2020inceptiontime,rank1}, we utilized a standard setting that includes unified hyper-parameters across all datasets. A uniform set of hyper-parameters provides a fair and comparable testing ground. Batch size is the only parameter that does not have an identical value across all datasets. Batch size is set to a lower number for some datasets due to hardware limitations (EigenWorms: 6, EthanolConcentration: 16, MotorImagery: 4, SelfRegulationSCP2: 32). For the rest of the Datasets, Batch size is set to 64. The rest of the training parameters for all models and all datasets are the same. The initial learning rate, loss function, optimizer, and the number of epochs are set to $0.0001$, categorical cross-entropy, Adam, and $400$, respectively. We also used a learning rate scheduler, which decreases the learning rate by a factor of 0.1 if the validation loss does not improve after 20 consecutive epochs. We used their exact model specifications, including kernel sizes and numbers, for the CNN-based models (FCN, ResNet, and InceptionTime). For the self-attention encoders, the number of layers and heads are set to 1. Moreover, the input embedding dimension size is set to 128. Our experiments are performed on a Compute Canada~\cite{compute-canada} node equipped with an NVIDIA V100 Volta GPU (32G HBM2 memory) unit.

\subsection{Comparison with state of the art}
In this section, we compare five state-of-the-art multivariate TSC models (\cite{chen_da-net_2022,tapnet,tang_omni-scale_2022,xcm,shapenet}) against three baseline deep learning TSC models (FCN~\cite{wang2017time}, ResNet~\cite{wang2017time}, and InceptionTime~\cite{fawaz2020inceptiontime}). Table~\ref{tb:all} encapsulates the performance accuracies for these methods. As the average accuracies on UEA show, all five state-of-the-art models~\cite{chen_da-net_2022,tapnet,tang_omni-scale_2022,xcm,shapenet} underperform compared to the three baseline models. Therefore, we chose to add the proposed GTA and TPS augmentations to the baseline models. In Table~\ref{tb:all}, we also compared the baseline models (FCN, ResNet, and InceptionTime) with their GTA- or TPS-augmented counterparts. For GTA augmentation, the GTA block is added after each computing layer of the model, as shown in Fig.~\ref{fig:fcngta}. As for TPS, the modified model is shown in Fig.~\ref{fig:st}. 

The accuracy values for UEA datasets are shown in Table~\ref{tb:all}. These results are the average   of 5 independent runs for each model. 
Rank Average (the mean rank of each model in terms of its accuracy for each specific dataset, lower numbers are better) is presented in the last row of Table~\ref{tb:all}. Bold numbers indicate the highest value in each row. Colored numbers in each vertical section highlight results higher than the base model (shown in red).

From Table~\ref{tb:all}, ``InceptionTime+ TPS" delivers the highest average accuracy. Adding TPS consistently improves the base models' accuracy by up to 3.0\%. Adding GTA also improves the accuracy of all four models, though only marginally for the model with the largest temporal receptive field (InceptionTime). From the rank average metric, the highest-ranking models are ``InceptionTime + TPS" and ``InceptionTime + GTA". Adding PE to TPS only improves accuracy over TPS for the FCN base model, demonstrating that the effectiveness of Absolute Positional Encoding (PE) depends on the base model's architecture. 

\begin{figure}[!t]
\centering
\includegraphics[width=0.49\textwidth]{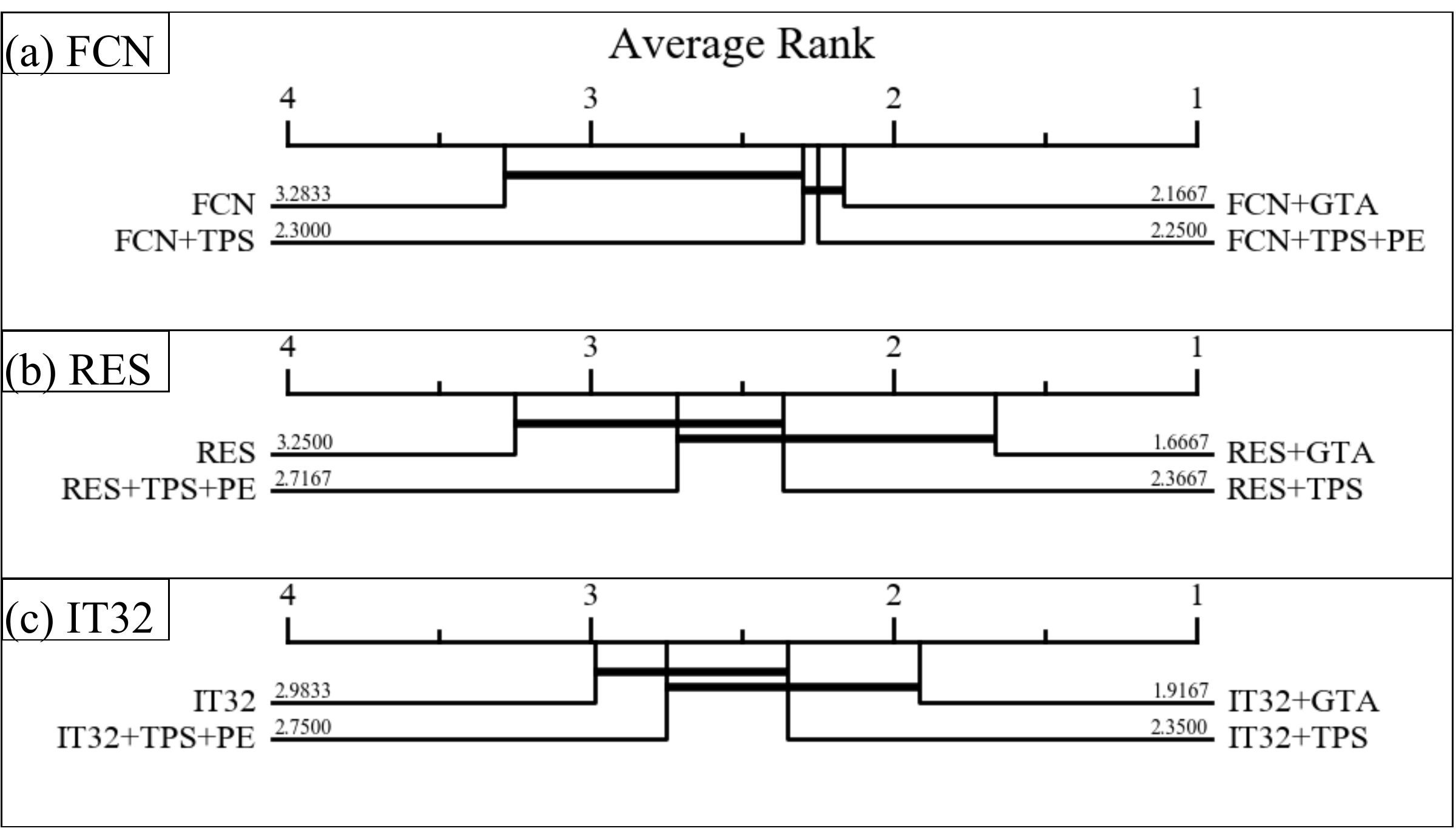}
\caption{CD diagram comparing the average ranks of different networks which use the same base models of (a) FCN, (b) ResNet, (c) and IT.}
\label{fig:cd}
\end{figure}
Figure~\ref{fig:cd} depicts the Wilcoxon-Holm post hoc test Critical Difference (CD) diagrams for each base model section of Table~\ref{tb:all} (one for every four columns under each base model). IT32 and RES stand for InceptionTime and ResNet, respectively. The CD diagrams imply that GTA model is producing better rankings than TPS model. However, there are no significant probabilistic differences between the two.

\subsection{Standalone TPS performance analysis}
\begin{table}[!t]
\setlength{\tabcolsep}{3pt}

  \centering
  \caption{Self-attention and TPS accuracy [\%] comparison on UEA datasets.}
\begin{threeparttable}
\begin{tabular}{l|c|c|c|c|c|c|c|c}
\toprule
Model & \multicolumn{1}{c}{-} & - & TUPE & DeBERTa & SA & TPS & SA+ & TPS+  \\
  & \multicolumn{1}{c}{} &   & \cite{ke_rethinking_2021} & \cite{he2021deberta} &   &   & PE & PE \\
\midrule
Inj.M & \multicolumn{1}{c}{-} & - & MAM & Both & None & MAM & APE & Both \\
Ref.P & \multicolumn{1}{c}{-} & - & Both & Both & None & Rel & Abs & Both \\
\midrule
Dataset & D\tnote{1} & N\tnote{2} & \multicolumn{1}{c}{-} & - & \multicolumn{1}{c}{-} & \multicolumn{1}{c}{-} & \multicolumn{1}{c}{-} & - \\
\midrule
AWR & 9 & 144 & 76.7 & \textbf{96.7} & 79.2 & 91.7 & 87.4 & 94.3 \\
AF & 2 & 640 & 42.7 & 40.0 & 38.7 & \textbf{53.3} & 36.0 & 46.7 \\
BM & 6 & 100 & \textbf{100.0} & \textbf{100.0} & 96.4 & \textbf{100.0} & 99.0 & \textbf{100.0} \\
CT & 3 & 182 & 97.3 & 97.0 & 86.1 & 97.2 & 97.0 & \textbf{98.5} \\
CR & 6 & 1197 & 93.4 & 98.1 & 95.4 & \textbf{98.6} & 91.7 & 98.6 \\
DDG & 1345 & 270 & 66.8 & 63.6 & 64.9 & 70.0 & 70.0 & \textbf{74.0} \\
EW & 6 & 17984 & 82.4 & \textbf{86.3} & 83.7 & 85.5 & 75.2 & 82.4 \\
EP & 3 & 206 & 86.4 & 90.4 & 79.7 & \textbf{96.4} & 82.3 & 95.7 \\
EC & 4 & 65 & 80.4 & 80.7 & 57.4 & 75.9 & 79.6 & \textbf{81.5} \\
ER & 3 & 1751 & 37.4 & 32.6 & 33.5 & \textbf{39.9} & 34.4 & 36.1 \\
FD & 144 & 62 & 58.8 & 55.3 & 54.4 & 56.4 & 58.9 & \textbf{62.3} \\
FM & 28 & 50 & 55.0 & 57.2 & 53.8 & \textbf{58.0} & 53.8 & \textbf{58.0} \\
HMD & 10 & 400 & 39.7 & 38.4 & 37.4 & 43.2 & 37.0 & \textbf{45.9} \\
HW & 3 & 152 & \textbf{15.4} & 7.4 & 9.4 & 14.9 & 11.2 & 11.1 \\
HB & 61 & 405 & 76.4 & 78.0 & 75.7 & \textbf{78.5} & 75.3 & 77.1 \\
IW & 200 & 30 & N/A & 64.9 & 55.7 & 65.2 & 64.3 & \textbf{65.2} \\
JV & 12 & 29 & 97.0 & 97.5 & 82.3 & 97.8 & 96.9 & \textbf{98.9} \\
LIB & 2 & 45 & 29.6 & 13.6 & 21.4 & 43.3 & 51.4 & \textbf{58.9} \\
LSST & 6 & 36 & 66.7 & 68.2 & 61.3 & \textbf{69.4} & 63.5 & 67.9 \\
MI & 64 & 3000 & 60.8 & 61.8 & 59.1 & \textbf{65.0} & 55.6 & 63.0 \\
NA & 24 & 51 & 87.3 & 84.8 & 74.9 & 87.2 & 87.8 & \textbf{95.6} \\
PEMS & 963 & 144 & 80.6 & 79.1 & 67.9 & 81.5 & 79.9 & \textbf{83.8} \\
PD & 2 & 8 & N/A & 87.4 & 73.0 & 95.3 & 95.9 & \textbf{97.1} \\
PM & 11 & 217 & 13.5 & 8.3 & 7.8 & \textbf{16.5} & 9.2 & 14.2 \\
RS & 6 & 30 & 81.8 & 82.8 & 73.5 & 83.6 & 81.4 & \textbf{84.9} \\
SRS1 & 6 & 896 & 85.2 & 86.2 & 83.4 & 83.3 & 84.0 & \textbf{88.4} \\
SRS2 & 7 & 1152 & 58.9 & 58.6 & 56.1 & \textbf{62.2} & 58.0 & 62.2 \\
SAD & 13 & 93 & 97.1 & 95.6 & 96.8 & 96.4 & 98.0 & \textbf{98.9} \\
SWJ & 4 & 2500 & 41.3 & 38.7 & 48.7 & 60.0 & 41.3 & \textbf{66.7} \\
UW & 3 & 315 & 69.7 & 47.9 & 42.6 & 46.9 & 60.0 & \textbf{71.9} \\
\midrule
Average &   &   & 67.1 & 66.6 & 61.7 & 70.4 & 67.2 & \textbf{72.7} \\
\midrule
Rank Avg &   &   & 3.3 & 3.8 & 5.2 & 2.3 & 4.2 & \textbf{1.7} \\
\bottomrule
\end{tabular}%

\begin{tablenotes}
\item[1] Dimension \item[2] Max Series Length
\end{tablenotes}
\end{threeparttable}

  \label{tb:bvgb}%
\end{table}%
This section presents experimental results and highlights the effectiveness of reformulating self-attention with Temporal Pseudo-gaussian augmentation in the temporal attention blocks. These experiments also include the accuracy comparison between TPS and two of the latest state-of-the-art works in positional information injection into transformers (TUPE~\cite{ke_rethinking_2021} and DeBERTa~\cite{he2021deberta}). DA-transformer~\cite{wu_da-transformer_2021} was also taken into account, but unlike the other two, there was no public reproduction material for DA-transformer, so it was left out of the quantitative comparison.

The baseline self-attention (SA) classifier is comprised of a TSC base model with an encoder using self-attention~\cite{vaswani2017attention} added at the output (this can be imagined as a modified version of what is shown in Fig.~\ref{fig:st}, where the PE is removed and SA replaces the TPS encoder). The addition of PE to each of SA and TPS models creates SA+PE and TPS+PE, respectively. The TSC base model here (Fig.~\ref{fig:st}) is an FC layer that converts the data's dimension to 128 for input to the self-attention and TPS encoders.

The performance accuracy for each model on UEA datasets is presented in Table~\ref{tb:bvgb}. The numbers in each row represent the average of five runs. Bold numbers highlight the highest value in each row. From this table, replacing SA with the TPS block improves the accuracy of the network by an average of 8.7\%. APE + SA improves accuracy by 5.5\%. TPS + PE results in the highest accuracy increase of 11\%.

TUPE and DeBERTa do not reach the average accuracy of the SA+PE model, even though they both have newer ways of processing positional information. That could be because each method was made to deal with different problems in different Natural Language Processing (NLP) tasks. Therefore, they lost their generality in comparison to the original SA encoder. As a result, a lower average accuracy than TPS was expected. Even though DeBERTA uses Rel and Abs reference points and APE and MAM injection methods, its average accuracy is still lower than TPS + PE. Based on the characteristics of TUPE, one would expect a better performance than TPS and worse than TPS+PE. It, however, did not catch up to either of them. A limitation of TUPE algorithm is that it cannot operate on short sequences. As a result, it could not generate any results for InsectWingbeat and PenDigits datasets.

Comparison of the accuracy results for FCN and ResNet in Table~\ref{tb:all} with the standalone TPS + PE unit shown in Table~\ref{tb:bvgb} implies that the standalone TPS + PE unit performs better than both FCN and ResNet. However, TPS has fewer learnable parameters and fewer computational complexities than FCN and ResNet. The number of learnable parameters for TPS is:
\begin{equation} \label{eq:noptp}
\begin{split}
No\ Parameters=&\left(l+9+\frac{l}{h}\right)\times d^2\\
& +\left(d_{dataset}+\ 2l\ +11\right)d
\end{split}
\end{equation}

In this equation, $l$ is the number of layers (1), $h$ is the number of attention heads (1), $d$ is the hidden dimension size (128), and $d_{dataset}$ is the dimension of the test dataset. Based on these values, the number of learnable parameters for TPS standalone model is about $d_{dataset} \times 128 + 182k$. Meanwhile, the number of learnable parameters for the FCN model is $(8\times d_{dataset}+1)\times 128 + 267k$, and ResNet is almost twice that. The learnable PE unit also adds additional learnable parameters of $ 2N \times d_{dataset}$ ($N$ is the max sequence length). Based on Table~\ref{tb:bvgb}, TPS+ PE has more learnable parameters than FCN only for ``EigenWorms” and ``MotorImagery” datasets. Interestingly, for both cases, standalone TPS without PE performs better than both TPS+PE and all base models. Additionally, if the number of learnable parameters is a limiting factor for PE, using non-learnable PE functions~\cite{ vaswani2017attention} could be explored as an alternative.

\subsection{Qualitative Attention Analysis}
\begin{figure}[!t]
\centering
\includegraphics[width=0.49\textwidth]{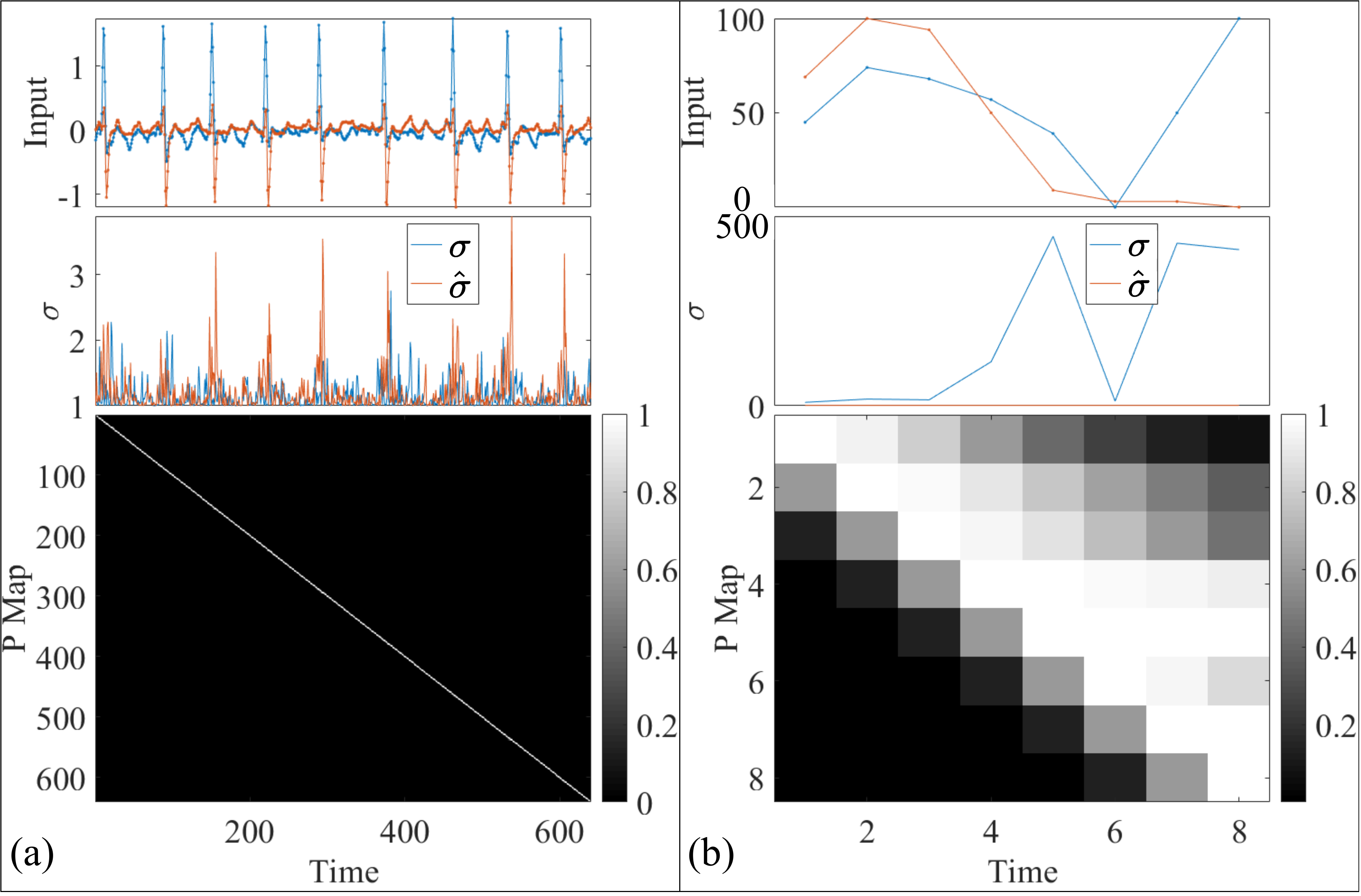}
\caption{$P$ and $\sigma$ comparison for samples from (a) AtrialFibrillation, (b) PenDigits datasets. Top) 2D input, Middle) $\hat{\sigma}$ and $\sigma$ as the backward and forward Gaussian neighbor attention Std, Bottom) $A_2^T$ as defined in Eq.~\eqref{eq:tps2}.}
\label{fig:pmap}
\end{figure}

This section visualizes the effect of asymmetrical pseudo-Gaussian positional attention on the content-correlation attention matrix. Pseudo-Gaussian attention injects the positional information into the transformer and forces the encoder to find new and unexplored relations between the time samples. Visualizations are performed on two instances trained and tested on AtrialFibrillation~\cite{bagnall2018uea} and PenDigits~\cite{bagnall2018uea} datasets. Each dataset consists of 2D multivariate time series arrays. However, their series lengths are substantially different (640 for AtrialFibrillation and 8 for PenDigits). AtrialFibrillation is a dataset composed of two-channel ECG signal recordings. The task is to predict spontaneous atrial fibrillation (AF) termination (3 classes) from 5-second long recorded instances with a 128 samples per second sampling rate. PenDigits is a dataset composed of time recorded x and y coordinates of a pen-tip, while the pen is used to write down a digit (0-9) on a digital $500\times 500$ screen. The axial coordinates are normalized to $100\times 100$ and resampled into 8 time samples. Fig.~\ref{fig:pmap} shows the calculated $P$ and $\sigma$ values (Eq.~\eqref{eq:tps2}) from the TPS model for two multivariate time series samples from the above two datasets.

\begin{figure}[!t]
\centering
\includegraphics[width=0.49\textwidth]{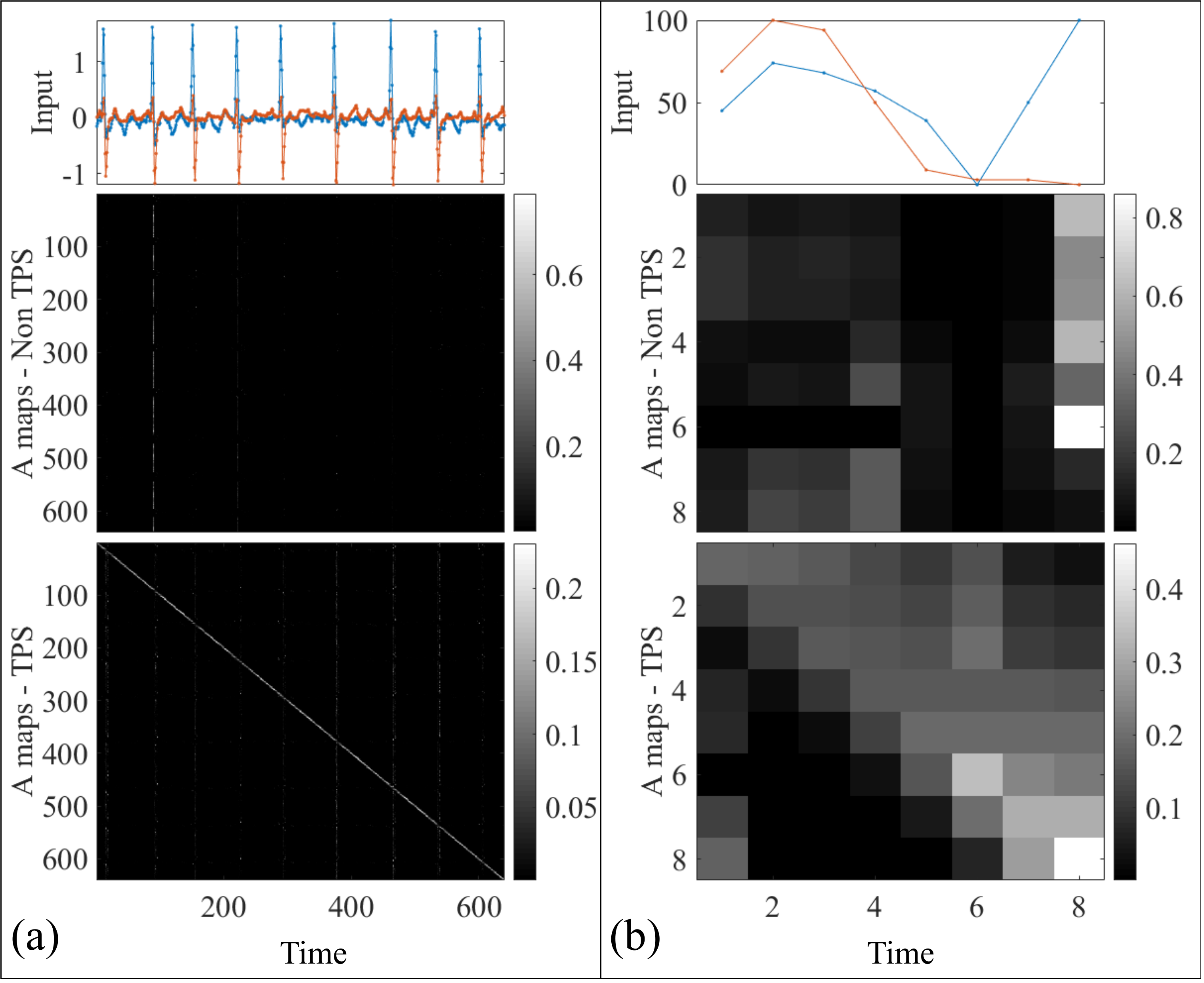}
\caption{Attention map comparison between TPS and SA models on samples from a) AtrialFibrillation, b) PenDigits datasets.}
\label{fig:amap}
\end{figure}
As shown in Fig.~\ref{fig:pmap}, separate calculations of $\sigma$ and $\hat{\sigma}$ provide flexibility in the distribution of neighboring attention. For AtrialFibrillation, the distribution of attention varies across different points. $\hat{\sigma}$ values seem higher than $\sigma$ values, indicating that more attention is placed on the previous samples. Also, the pseudo-Gaussian attention is only spanning across a small temporal range compared to the series length. For the PenDigits sample, the attention is directed toward future samples with larger $\sigma$ values that make it span across the entire series.

Fig.~\ref{fig:amap} shows the attention maps for SA and TPS models ($A$ in Eq.~\eqref{eq:tps}) for two sample multivariate time series. From the second row, we can conclude that the self-attention mechanism takes a weighted average of the key time points (possibly because of the GAP layer). However, for TPS, self-attention is forced to identify connections between multiple time data points and the data points along the diagonal direction. 

Our experimental results show that adding our proposed block to the existing TSC models can make them work better. Since there is no statistical difference between how well the two blocks improve performance, choosing the best block depends on the task and how easy it is to implement. Furthermore, it was shown that a TPS block could be used as a standalone TSC model with comparatively good performance and fewer computational complexities compared to the state-of-the-art. The asymmetrical pseudo-Gaussian positional attention is the main reason the TPS block works well. This is because it feeds relative positional information into the transformer model and forces the transformer to make new and better content-correlation attention matrices.

\section{Conclusions}
This paper presented two novel attention blocks, GTA and TPS, for deep learning-based TSC networks. We showed that incorporating these two blocks into DNN TSC models could improve their performances. GTA is proposed as a sublayer attention block, placed after each 1D Convolutional layer block. In contrast, TPS is presented as an add-on block that could reprocess the output of a TSC model. Experiments on UEA benchmark dataset archive highlighted the advantage of adding TPS and GTA blocks to three state-of-the-art baseline deep learning-based TSC models. These experiments demonstrated that both blocks could improve the accuracy and average rank in all three state-of-the-art. However, the improvement varies according to the application; in some cases, it is marginally and in others substantially better. Since there is no probabilistic difference between the two methods, the choice could be based on the task. We also showed that TPS block could be used as an independent TSC unit. The standalone TPS unit is better at TSC compared to the state-of-the-art in transformer’s positional information injection methods. Additionally, an independent TPS unit coupled with PE performed better than both base FCN and ResNet models with almost half and one-sixth of the number of learnable parameters, respectively.
\ifCLASSOPTIONcaptionsoff
  \newpage
\fi



%
\printbibliography









\end{document}